\documentclass{article}
\usepackage[a4paper, margin=2cm]{geometry}
\usepackage{graphicx, multirow, booktabs} 
\usepackage{hyperref, comment}
\usepackage{amsmath}
\usepackage{authblk} 
\title{\textbf{Deep Learning Characterizes Depression and Suicidal Ideation from Eye Movements}}

\author[1]{\normalsize Kleanthis Avramidis}
\author[2,3]{Woojae Jeong}
\author[2]{Aditya Kommineni}
\author[2]{Sudarsana R. Kadiri}
\author[1]{Marcus Ma}
\author[4]{Colin~McDaniel}
\author[4]{Myzelle Hughes}
\author[8]{Thomas McGee}
\author[6]{Elsi Kaiser}
\author[6]{Dani Byrd}
\author[4]{Assal Habibi}
\author[5]{B.~Rael~Cahn}
\author[8]{Idan~A.~Blank}
\author[1]{Kristina Lerman}
\author[2]{Takfarinas Medani}
\author[2,3]{Richard M. Leahy}
\author[1,2,6,7]{Shrikanth~Narayanan}

\affil[1]{\normalsize Thomas Lord Department of Computer Science, University of Southern California}
\affil[2]{\normalsize Ming Hsieh Department of Electrical and Computer Engineering, University of Southern California}
\affil[3]{\normalsize Alfred E. Mann Department of Biomedical Engineering, University of Southern California}
\affil[4]{\normalsize Brain and Creativity Institute, University of Southern California}
\affil[5]{Department of Psychiatry and Behavioral Sciences, University of Southern California}
\affil[6]{\normalsize Department of Linguistics, University of Southern California}
\affil[7]{\normalsize Department of Psychology, University of Southern California}
\affil[8]{\normalsize Department of Psychology, University of California, Los Angeles}

\begin{document}
\maketitle

\begin{abstract}
Identifying physiological and behavioral markers for mental health conditions is a longstanding challenge in psychiatry. Depression and suicidal ideation, in particular, lack objective biomarkers, with screening and diagnosis primarily relying on self-reports and clinical interviews. Here, we investigate eye tracking as a potential marker modality for screening purposes. Eye movements are directly modulated by neuronal networks and have been associated with attentional and mood-related patterns; however, their predictive value for depression and suicidality remains unclear. We recorded eye-tracking sequences from 126 young adults as they read and responded to affective sentences, and subsequently developed a deep learning framework to predict their clinical status. The proposed model included separate branches for trials of positive and negative sentiment, and used 2D time-series representations to account for both intra-trial and inter-trial variations. We were able to identify depression and suicidal ideation with an area under the receiver operating curve (AUC) of 0.793 (95\% CI: 0.765–0.819) against healthy controls, and suicidality specifically with 0.826 AUC (95\% CI: 0.797–0.852). The model also exhibited moderate, yet significant, accuracy in differentiating depressed from suicidal participants, with 0.609 AUC (95\% CI 0.571--0.646). Discriminative patterns emerge more strongly when assessing the data relative to response generation than relative to the onset time of the final word of the sentences. The most pronounced effects were observed for negative-sentiment sentences, that are congruent to depressed and suicidal participants. Our findings highlight eye tracking as an objective tool for mental health assessment and underscore the modulatory impact of emotional stimuli on cognitive processes affecting oculomotor control.
\end{abstract}
\section*{Introduction}

Depression is one of the most significant and prevalent mental health disorders, affecting approximately 280 million people globally, according to the World Health Organization~\cite{IHME_GBD_2023}. Suicidal ideation, or suicidality, represents another critical public health challenge that is frequently comorbid with depression, with over 700,000 annual deaths attributed to suicide worldwide~\cite{WHO_suicide}. Extensive research has focused on identifying biomarkers and behavioral indicators for early screening of these conditions, drawing from clinical evaluations, behavioral observations (e.g., social withdrawal, sleep disturbance), and environmental factors (e.g., socioeconomic stress). However, detecting suicidality remains particularly challenging due to the stigma surrounding mental health, the covert nature of symptoms, and the frequent overlap of risk factors with psychiatric comorbidities.

Identifying objective physiological and behavioral markers for mental health disorders is thus an important research direction. While there are established clinical procedures for assessing and measuring physical health, the domain of mental health is still lacking well-established biomarkers~\cite{abi2023candidate}, and assessment primarily relies on clinical evaluation through direct observation, self-reports, or interviews~\cite{bone2017signal}. This methodology is limiting, because such measures can be influenced by recall bias, social desirability, and variability in self-awareness~\cite{harari2017smartphone,baumeister2007psychology}, making them less reliable for objective assessment.
Recent advances in sensing technologies and computational analysis of neural, physiological, and behavioral signals are beginning to provide opportunities for more precise and nuanced access to human experiences that were previously difficult to quantify~\cite{hauser2022promise, hansen2023speech}.

A significant portion of neurophysiological research aims to uncover objective markers of depression and suicidal ideation by analyzing brain and body responses to appropriately designed stimuli. Electroencephalography (EEG) and Magnetic Resonance Imaging (MRI) are widely used to investigate neural correlates~\cite{schmaal2020enigma,de2019depression,dennis2010frontal,murray2011localization,botteron2002volumetric}. 
Despite these advances, no single measure is considered a definitive diagnostic or prognostic marker, and therefore none can be
straightforwardly used on its own in a clinical setting. 
In this context, eye tracking is a promising yet underexplored tool, and eye movements are increasingly recognized in mental-health research as reflections of cognitive and emotional processes~\cite{skaramagkas2021review}. For instance, studies show that gaze patterns during emotional stimuli (e.g., faces, scenes) correlate with depression severity, such that individuals with depressive symptoms often exhibit attentional biases—such as prolonged fixation on negative cues~\cite{gotlib2004attentional} or avoidance of positive ones~\cite{suslow2001detection}—or reduced diversity in their patterns of visual exploration. Moreover, because eye movements are difficult to consciously control, eye tracking may provide an objective and accessible screening tool for a wide range of populations. 
Despite this potential, eye-tracking analyses have been constrained by reliance on handcrafted features~\cite{skaramagkas2021review}. While these metrics offer insights into overt attentional behavior, they often fall short of capturing the dynamic spatiotemporal patterns embedded in raw gaze data~\cite{huettig2011using}. More recently, deep learning on eye-tracking data has been explored to uncover such markers of dementia~\cite{sriram2023classification} and also depression~\cite{kobo2024classification} with improved performance compared to handcrafted features. This tendency underscores the potential of adopting more advanced computational approaches for behavior analysis.

Our study develops a novel deep learning framework to capture the temporal dynamics of short, event-based eye movements during an affectively loaded reading task. The proposed model is suitable for a wide range of experimental designs in neuroscience and psychology~\cite{hodgson2009saccadic,hollingworth2020eye}, particularly when considering multi-trial experimental paradigms. Our pipeline uses bootstrap resampling of the experimental trials (Figure~\ref{fig:methods}E) to regularize model training and improve generalization to novel experimental trial sets. We applied this framework to investigate potential markers of depression and suicidality in eye-movement data from a cohort of 126 young adults identified as either healthy controls (C), depressed (D) or suicidal (S) during the recruitment phase. All subjects undertook a sentence reading task during which they read 160 different sentences varying in affective content, presented word by word (Figure~\ref{fig:methods}A), and indicated whether they agree or not with each sentence. The proposed method provides insights into the timing of differential responses related to depression and suicidality as well as the relevance of the stimulus valence and association to self-reported behavioral attributes.
\section*{Results}

\subsection*{Experimental Setup}

\begin{figure*}[t]
    \centering
    \includegraphics[width=\linewidth]{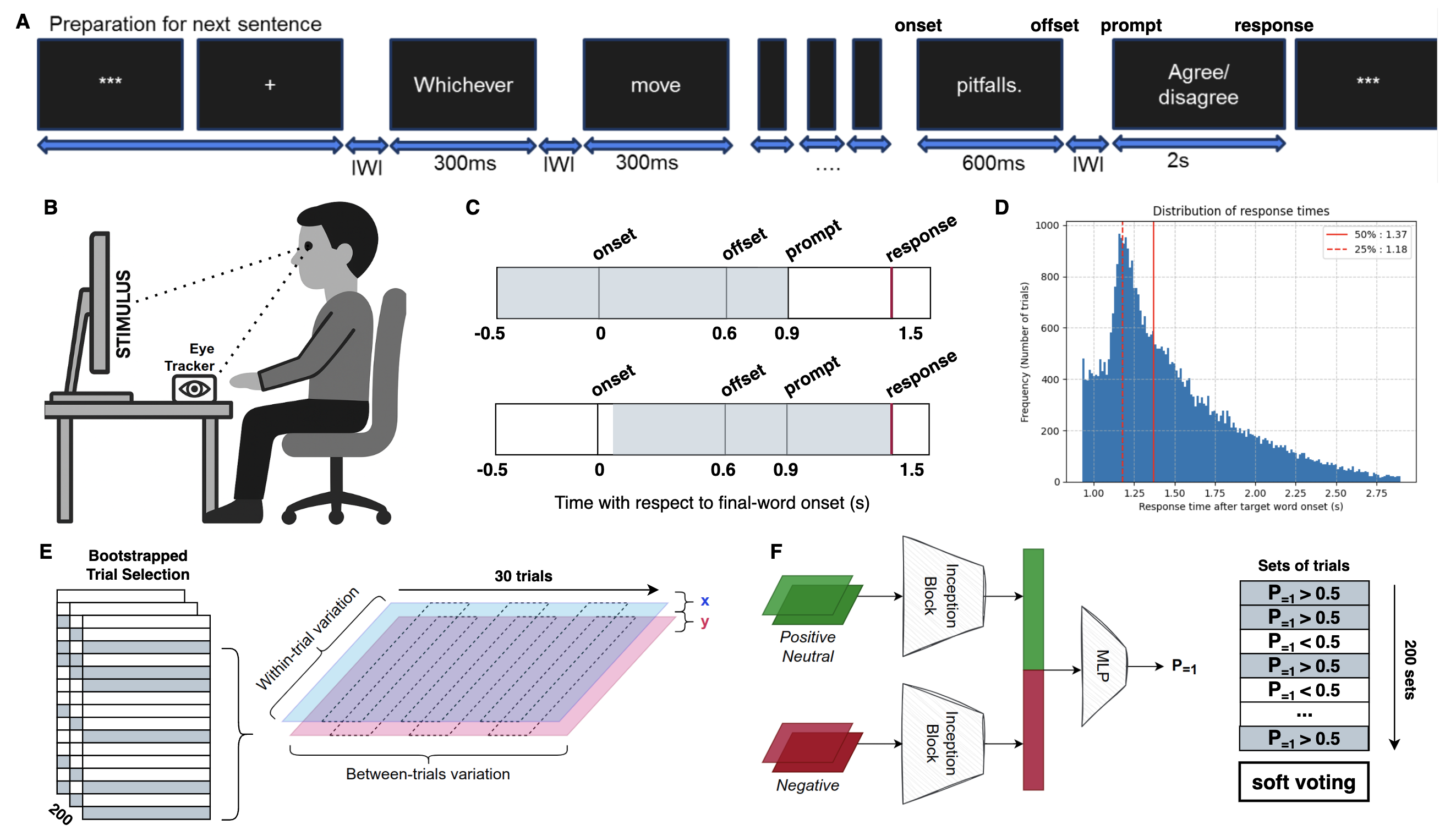}
    \vspace{-0.6cm}
    \caption{\footnotesize\textbf{Experimental setup and deep learning pipeline:} (\textbf{A}) The stimulus sequence presentation for a single trial of the sentence viewing task. (\textbf{B}) Schematic overview of the data collection setup, with only the eye-tracking setup depicted. (\textbf{C}) The two different strategies for segmenting eye-tracking trials. Top: Signals aligned to onset of the final word. Bottom: Signals aligned to the subject's response and reaching back to 1200\,ms before the response. (\textbf{D}) Histogram distribution of response times with respect to final-word onset, for the entire dataset. Highlighted are the median and mode of the distribution, which determined the trial segmentation. (\textbf{E}) The process of creating input samples by bootstrapping 200 sets of 30 trials per participant. Each set is input to the deep learning model as a 2-channel 2D input, reflecting both inter-trial and intra-trial variability. Channels correspond to horizontal (x) and vertical (y) movements. (\textbf{F}) The 2D input is processed by an Inception~\cite{inception} (convolutional) block and then concatenated before the classifier. MLP stands for multi-layer perceptron. $P_{=1}$ is the output probability of the sample belonging to the positive (non-control) class. The final subject-level prediction is derived by the proportion of input trial sets scoring $P_{=1}>0.5$.}
    \label{fig:methods}
\end{figure*}

A total of approximately 37,000 eye movement trials from 126 participants during sentence reading were used to train and validate the deep learning model within a nested cross-validation setup. To that end, we split the participants into 5 distinct folds that were approximately equal in size. We modeled the three available groups in multiple binary classification scenarios, primarily focusing on distinguishing clinical participants (groups D+S, henceforth DS) from healthy controls (group C). We explored two distinct input schemes (Figure~\ref{fig:methods}C): one centered on the final word of the sentence, which conveyed the affective content, henceforth \textit{reading}-model, and another centered on the participant’s subsequent response, indicating whether they agreed or disagreed with the presented sentence, henceforth \textit{response}-model. The former includes eye-movement data starting 500\,ms before the final word appeared on the screen, and
continuing for 900\,ms to capture immediate gaze dynamics upon encountering emotionally salient stimuli, which are crucial for identifying cognitive biases. In contrast, the \textit{response}-model includes eye movement from 1200\,ms before and up until the moment of response, hence captures patterns associated with attentional processing leading to decision making; the temporal evolution of eye movement signals in this case is time-locked to the button press related to this decision making.

\subsection*{Model Performance}

\begin{table}[ht]
\centering
\begin{tabular}{@{}lcccccccc@{}}
\toprule
\multirow{2}{*}{Scenario} & \multirow{2}{*}{\#C} & \multirow{2}{*}{\#D} & \multirow{2}{*}{\#S} & $P_C$ & $P_D$ & $P_S$ & AUC & $p$-value \\
& & & & (95\% CI) & (95\% CI) & (95\% CI) & (95\% CI) & \\
\midrule
\multicolumn{8}{@{}l}{\textit{Response-model performance}} \\
\addlinespace
C vs. DS & 43 & 40 & 43 & 
0.365 & 0.699 & 0.785 & 0.793 & $< 0.001$ \\
& & & & (0.332--0.396) & (0.669--0.727) & (0.760--0.810) & (0.765--0.819) & \\
\addlinespace
C vs. D & 43 & 40 & -- & 
0.352 & 0.662 & \textbf{0.717} & 0.740 & $< 0.001$ \\
& & & & (0.321--0.387) & (0.631--0.694) & \textbf{(0.689--0.746)} & (0.705--0.771) & \\
\addlinespace
C vs. S & 43 & -- & 43 & 
0.326 & \textbf{0.546} & 0.742 & 0.815 & $< 0.001$ \\
& & & & (0.294--0.358) & \textbf{(0.511--0.578)} & (0.714--0.769) & (0.786--0.842) & \\
\addlinespace
D vs. S & -- & 40 & 43 & 
\textbf{0.362} & 0.437 & 0.545 & 0.589 & $0.002$ \\
& & & & \textbf{(0.324--0.399)} & (0.403--0.469) & (0.511--0.580) & (0.551--0.628) & \\
\addlinespace
\midrule
\multicolumn{8}{@{}l}{\textit{Reading-model performance}} \\
\addlinespace
C vs. DS & 43 & 40 & 43 & 
0.464 & 0.594 & 0.665 & 0.631 & $< 0.001$ \\
& & & & (0.431 -- 0.498) & (0.562 -- 0.623) & (0.633 -- 0.698) & (0.596 -- 0.661) & \\
\addlinespace
C vs. D & 43 & 40 & -- & 
0.400 & 0.499 & \textbf{0.586} & 0.586 & $0.024$ \\
& & & & (0.368--0.432) & (0.464--0.535) & \textbf{(0.551--0.619)} & (0.548 -- 0.626) & \\
\addlinespace
C vs. S & 43 & -- & 43 & 
0.414 & \textbf{0.567} & 0.579 & 0.627 & $< 0.001$ \\
& & & & (0.377--0.453) & \textbf{(0.533--0.602)} & (0.542--0.616) & (0.591--0.663) & \\
\addlinespace
D vs. S & -- & 40 & 43 & 
\textbf{0.517} & 0.465 & 0.563 & 0.577 & $0.034$ \\
& & & & \textbf{(0.476--0.553)} & (0.426--0.504) & (0.527--0.595) & (0.539--0.613) & \\
\bottomrule
\end{tabular}
\caption{\footnotesize\textbf{Performance overview of the deep-learning model:} Model variants were trained on 2 distinct time intervals and 4 classification tasks. Confidence intervals are computed via bootstrap resampling across 10 random seeds and all participants of the respective scenario. $P_i$ is the model-assigned probability to samples from group $i \in \{C, D, S\}$. To determine statistical significance for our results, we repeat the whole training process using permuted labels for training. We then bootstrap 1000 sets of 126 scores and count how many of those result in AUC higher than the bottom CI threshold of the respective classifier. Results in \textbf{bold} indicate zero-shot estimation of groups that were unseen in training. AUC is reported for the task optimized during training.}
\label{tab:classification_performance}
\end{table}

A summary of our model results is presented in Table~\ref{tab:classification_performance}. The \textit{response}-model for the main classification task (CvDS) achieved an AUC of 0.793 (95\% CI: 0.765--0.819, permutation test $p_\text{perm} < 0.001$). Control subjects received an average score of 0.365 (95\% CI: 0.332--0.396), whereas depressed and suicidal subjects scored 0.700 (95\% CI: 0.669--0.727) and 0.785 (95\% CI: 0.765--0.810), respectively (Figure~\ref{fig:roc-plots-resp}B-left). Based on those probability scores, the model differentiated the S group from controls with an AUC of 0.826 (95\% CI: 0.797--0.852), whereas the differentiation of the D group from controls showed an AUC of 0.759 (95\% CI 0.724--0.792). For the same experiment, the model yielded a sensitivity of 0.798 (95\% CI: 0.770--0.827) and a specificity of 0.674 (95\% CI: 0.628--0.716) when predicting C and DS at a probability threshold of $0.5$.  We note that the model was able to identify group S as having higher depression risk compared to group D, even though such information was not provided during model training. Nevertheless, the model differentiated group D from S with an AUC of 0.609 (95\% CI 0.571--0.646, $p_\text{perm} < 0.001$).

On the other hand, the \textit{reading}-model scored a moderate AUC of 0.631 (95\% CI: 0.596--0.661, $p_\text{perm} < 0.001$). Group-specific effects are compromised in this scenario (Figure~\ref{fig:plots-read}B-left), as control subjects scored a probability of 0.464 (95\% CI: 0.431--0.498), compared to 0.594 for depressed (95\% CI: 0.562--0.623) and 0.665 for suicidal (95\% CI: 0.633--0.698). Despite the smaller effect sizes, the discrimination ability of the classifier was still significantly higher than random chance, even at the lowest CI bound. The model differentiated the S group from controls with an AUC of 0.661 (95\% CI: 0.625--0.698), whereas the D group showed an AUC of 0.600 (95\% CI 0.563--0.640) against controls. Notably, the classifier can still, but weakly, differentiate between the two clinical groups D and S with an AUC of 0.583 (95\% CI 0.544--0.623, $p_\text{perm} = 0.022$). Overall, group differences are significantly more pronounced when analyzing the signal time-locked to the button press.

\subsection*{Pairwise Group Comparisons}

\begin{figure*}[t]
    \centering
    \includegraphics[width=\linewidth]{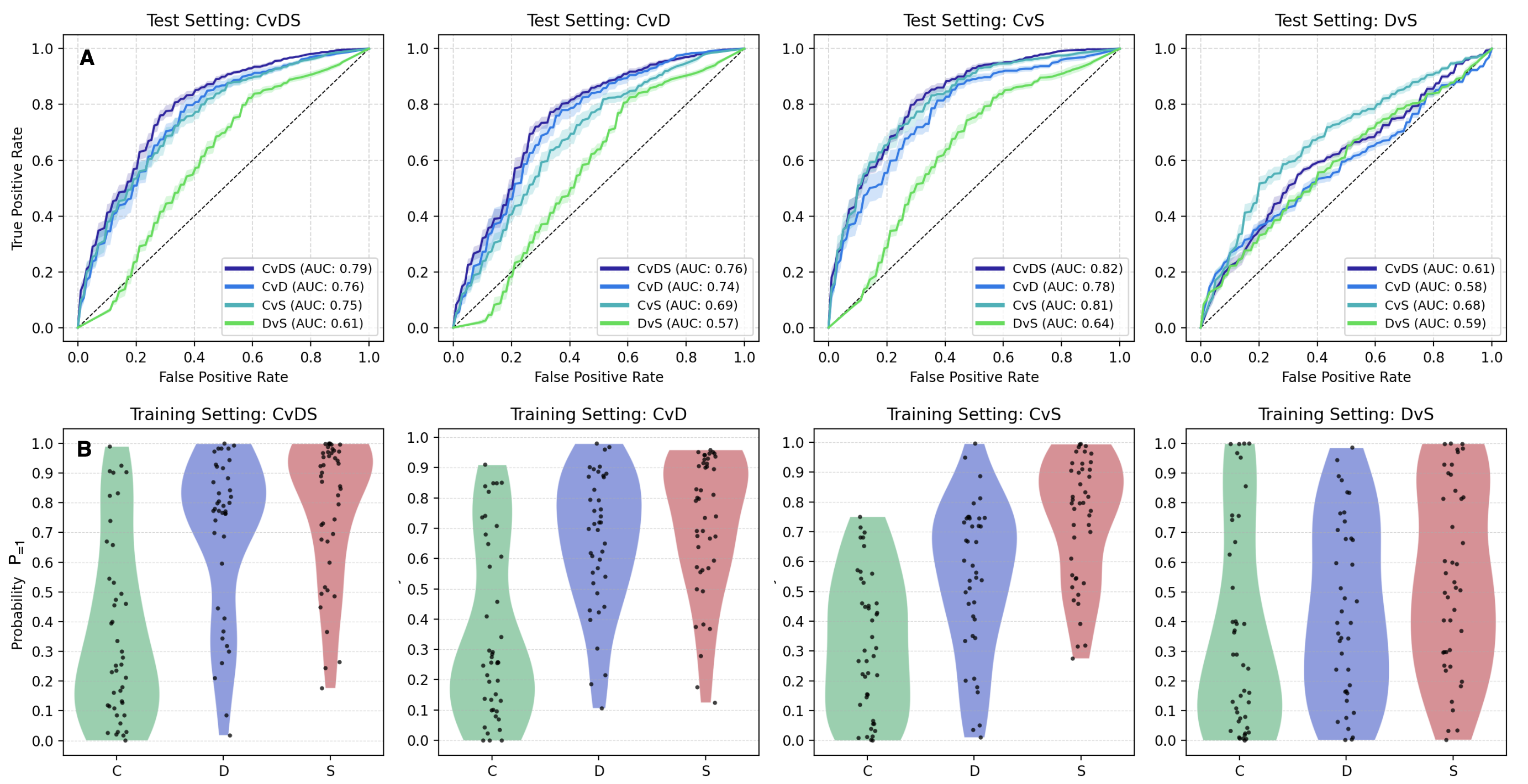}
    \vspace{-0.5cm}
    \caption{\footnotesize\textbf{Task-wise performance of \textit{response}-model:} (\textbf{A}) Receiver-operating curve (ROC) plots per binary test setting. Each plot compares the performance of four model variants trained on separate group setups, with AUC denoted in the respective legend. For example, the left-most figure depicts and compares the performance of four variants on the same evaluation task, i.e., CvDS. Each of those variants was trained on a different classification objective, as indicated by their naming. (\textbf{B}) Violin plots of group-wise model predictions per training configuration. Each participant is represented by a dot indicating the model-assigned probability of belonging to the positive class, averaged across 10 random seeds.}
    \label{fig:roc-plots-resp}
\end{figure*}

\begin{figure*}[h]
    \centering
    \includegraphics[width=\linewidth]{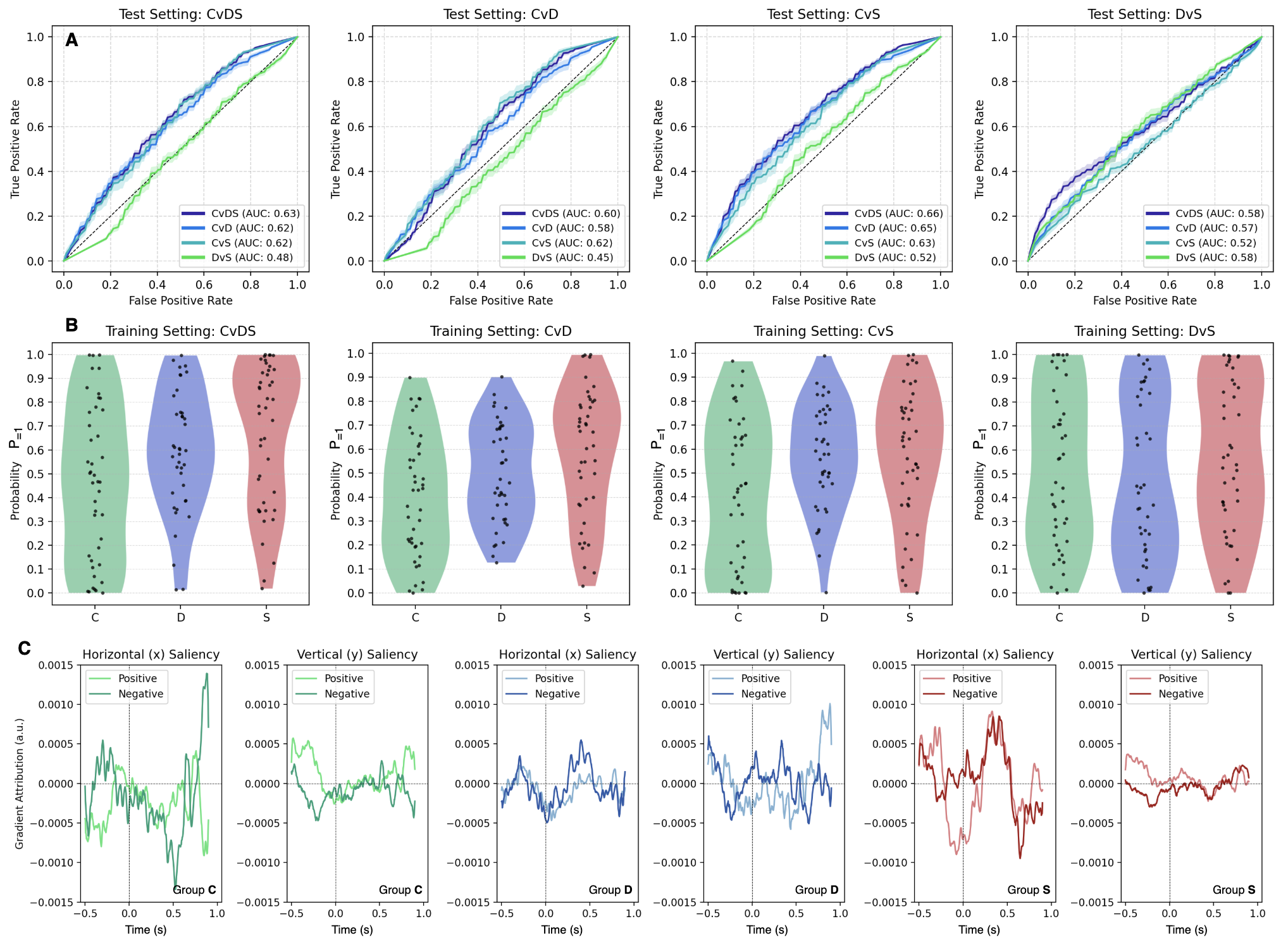}
    \vspace{-0.75cm}
    \caption{\footnotesize\textbf{Performance summary for the \textit{reading}-model:} (\textbf{A}) Receiver-operating curve (ROC) plots per binary test setting. Each plot compares the performance of 4 model variants trained on separate group setups, with AUC denoted in the respective legend. (\textbf{B}) Violin plots of group-wise model predictions per training configuration. Each participant is represented by a dot indicating the model-assigned probability of belonging to the positive class, averaged across 10 random seeds. (\textbf{C}) Time-resolved model attributions over raw gaze trials, computed with the Integrated Gradients [23] method. Positive scores correspond to higher attribution to the positive class, namely DS. Colors reflect participant groups: C: green, D: blue, S: red. Each time-series represents the average model attribution over all available samples from subjects of the respective group.}
    \label{fig:plots-read}
\end{figure*}

The remaining binary setups also demonstrate significant group separability, though they do not surpass the performance of the main CvDS training scenario (Figure~\ref{fig:roc-plots-resp}A). We attribute this difference in performance to the larger number of participants used to train the CvDS model. Nevertheless, we use the additional training setups to get an unbiased (zero-shot) estimation of groups that we kept unseen during training. The probabilities yielded for those groups are highlighted in bold in Table~\ref{tab:classification_performance}. In all cases, the \textit{response}-models successfully rank an unseen group correctly relative to the others, revealing robust indicators of depression severity, hypothesized schematically as C $<$ D $<$ S (Figure~\ref{fig:roc-plots-resp}B). Similar insights are deduced for the \textit{reading}-models (Figure~\ref{fig:plots-read}B), with the exception of the DvS setup, in which C is actually ranked higher than D.

When testing on the CvDS scenario, we found that training a \textit{response}-model to distinguish group C from D, or C from S, individually resulted in comparable performance with an AUC of 0.758 (95\% CI 0.729--0.787) and 0.752 (95\% CI 0.725--0.779), respectively (Figure~\ref{fig:roc-plots-resp}A-left). Both models were also tested for generalization on the respective unseen group (Figure~\ref{fig:roc-plots-resp}A, rest of columns). In specific, the CvS model underperformed the CvD one when tested on the latter configuration (two-sample t-test $p<0.001$), whereas the CvD model achieved an AUC of 0.776 (95\% CI 0.743--0.806) on the CvS task and was not significantly lower than the model trained on this task (two-sample t-test $p=0.211$). As seen in Figure~\ref{fig:roc-plots-resp}B-middle, this is because the CvD model achieves better containment of the C-group predictions, whereas the CvS model predicts suicidal individuals with higher confidence. Hence, when tested to differentiate the clinical groups D and S, the CvS model had the best performance with 0.681 AUC (95\% CI 0.645--0.718), indicating that the learning efficiency of the CvD task does not translate to better discrimination of the suicidal group. With respect to the onset of the final word, cross-group performance for the \textit{reading}-model was comparable in all cases (Figure~\ref{fig:plots-read}B).

\subsection*{Effect of Sentence Valence}

A key within-subject contrast in our experimental design involves the differential response to sentences with varying affective valence: those congruent with depression and suicidality (i.e., what we call negative) versus a controlled positive/neutral category. To quantify the predictive capability of each stimulus type, we trained separate classifiers using only one of the two conditions, i.e., either negative or positive/neutral, as well as a model with randomly shuffled condition labels (Figure~\ref{fig:interp}A-left). When trained exclusively on the negative sentences, the \textit{response}-model achieved an AUC of 0.757 (95\% CI: 0.729–0.784), trailing our main model by approximately 4\% on average (two-sample t-test $p = 0.039$). In contrast, the model trained on positive/neutral sentences yielded an AUC of 0.674 (95\% CI: 0.640–0.707), significantly lower than both the negative and the main model (two-sample t-test $p < 0.001$). These results highlight a stronger group-specific signal in responses to negative stimuli, consistent with known affective processing biases in depression and suicidality, and our hypothesis of a controlled positive/neutral category of stimuli. Nevertheless, the superior performance of the main model suggests an interaction between group membership and sentence valence. Accordingly, training the same model with shuffled sentiment labels yielded significant regressions (0.688 AUC, 95\% CI: 0.658–0.719).

\subsection*{Post-hoc Interpretability Analysis} 

\begin{figure*}[t]
    \centering
    \includegraphics[width=\linewidth]{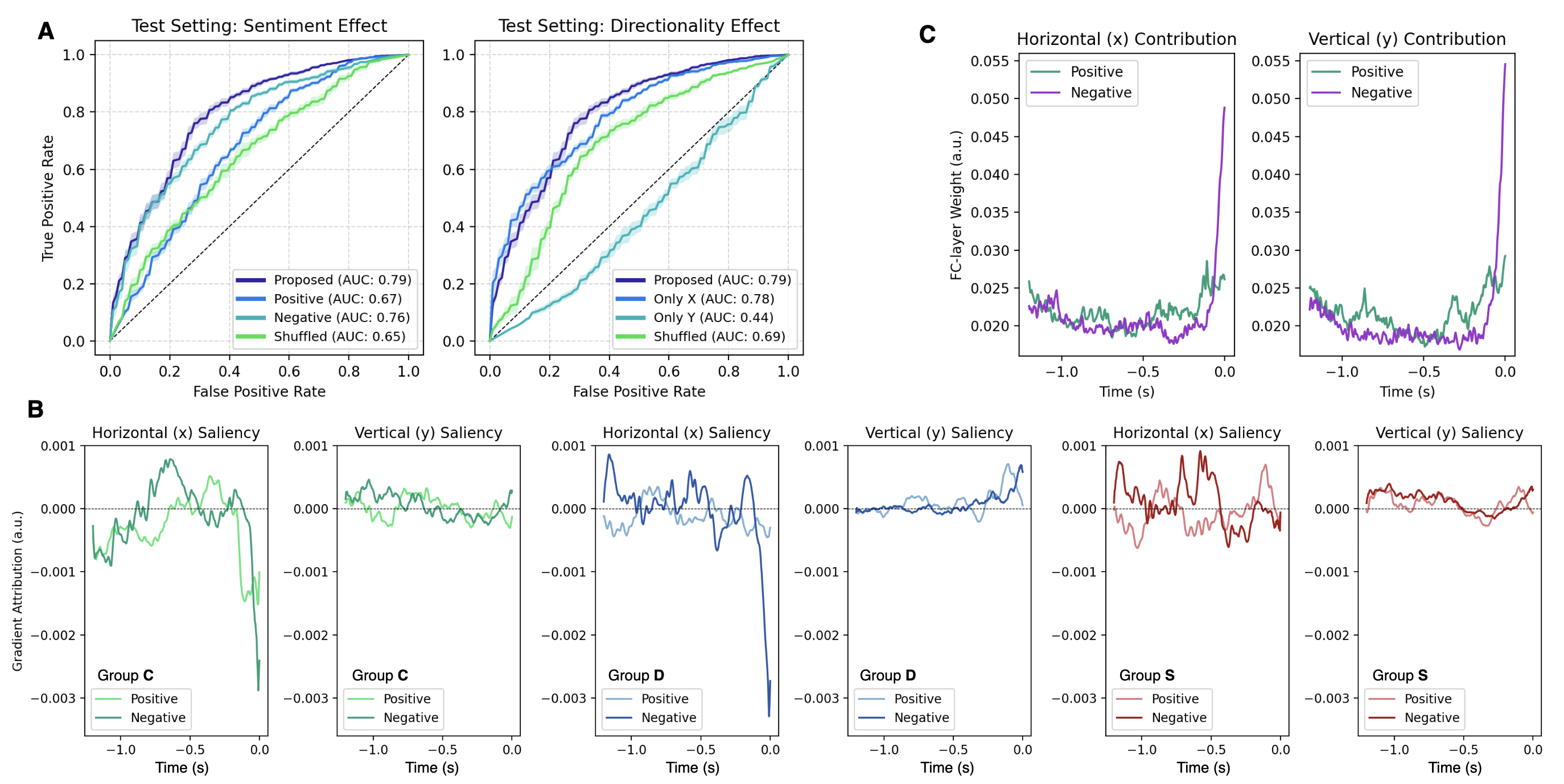}
    \vspace{-0.6cm}
    \caption{\footnotesize \textbf{Model interpretability and ablations:} Analysis is done here for the main \textit{response}-model, trained on the CvDS task. (\textbf{A}) Ablation study with respect to sentence sentiment (left) and input directionality (right). For each scenario, we train a control model by shuffling the attribute of interest. (\textbf{B}) Time-resolved model attributions over raw gaze trials, computed with the Integrated Gradients~\cite{sundararajan2017axiomatic} method. Positive scores correspond to higher attribution to the positive class, namely DS. Colors reflect participant groups: C: green, D: blue, S: red. Each time-series represents the average model attribution over all available samples from subjects of the respective group. (\textbf{C}) Model weights extracted from the first fully connected (FC) classification layer of the model. Based on the model architecture, this layer's input is a concatenated vector representation of both (x, y) directions and (positive, negative) conditions, thus letting us examine where the model attends to render the output predictions.}
    \label{fig:interp}
\end{figure*}

The deep learning performance provided evidence of information inherent to eye movements that differentiates depression and suicidality. However, while our experimental setup yields useful information on the timing and the valence of differentiable responses, it does not pinpoint specific behavioral differences between groups. To that end, we conduct a series of interpretability analyses, shown in Figure~\ref{fig:interp}.

Gradient attribution analysis (Figure~\ref{fig:interp}B) revealed that the model predominantly relies on horizontal eye movements during the final 200\,ms leading up to the subject's response. This trend was consistent across sentence valence and groups C, D, though it was particularly pronounced for negative stimuli, which elicited sharper saliency peaks that the model attributed to the control class (i.e., negative attribution). Group S exhibited more distributed effects that point to disengagement or less structured movement patterns. In contrast, vertical saliency remained minimal and flat, suggesting limited relevance of vertical gaze movements in driving model predictions. With respect to final-word onset (Figure~\ref{fig:plots-read}C), we also observe the strongest activations in the horizontal direction, peaking around 300\,ms after word onset for the clinical group, and close to the onset of the response prompt for the control group. Notably, the suicidal group displays a negative attribution near the final-word onset, potentially reflecting predictive processing based on the sentence context.

To complement those observations, we examined the magnitude of the weights of the \textit{response}-model’s first fully connected layer (Figure~\ref{fig:interp}C). The resulting temporal profiles confirmed that both horizontal and vertical components of negative stimuli received stronger weighting near the response time. While vertical movements were weighted heavily, that does not translate to differentiable responses in the saliency maps. This model behavior implies the presence of strong signal fluctuations that do not contribute to the training objective. Such movements can be associated with visually locating the off-screen response button, as the observed activations were not repeated in the readout of the \textit{reading}-model.

Finally, we performed ablative experiments to directly assess how sentiment and eye movement directionality contribute to classification performance (Figure~\ref{fig:interp}A-right). In the sentiment-based evaluation, the ablation study confirmed the diagnostic value of responses to negative content, as discussed also in the previous section. In the directionality-based ablation, the model trained with only horizontal (x) eye movements achieved nearly optimal performance with an AUC of 0.781 (95\% CI 0.756--0.808). In contrast, the vertical-movement-only model dropped below the random-chance level. These findings confirm that differentiable oculomotor dynamics are primarily expressed along the horizontal axis, especially during negative stimuli. Together, these interpretability results offer a mechanistic lens into the drivers of the observed diagnostic performance.

\subsection*{Sensitivity to Modeling Parameters}

\begin{figure*}[t]
    \centering
    \includegraphics[width=\linewidth]{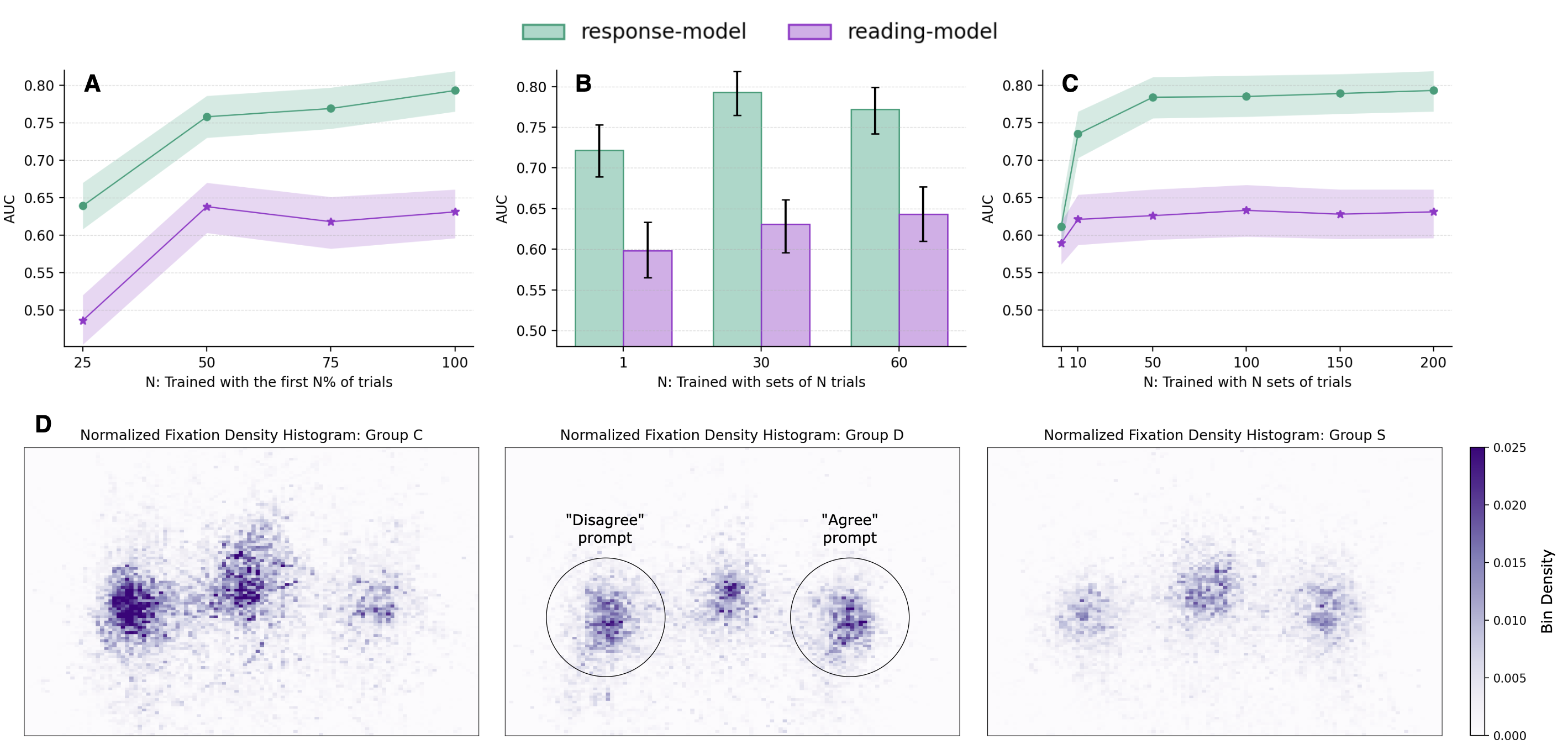}
    \vspace{-0.6cm}
    \caption{\footnotesize \textbf{Sensitivity to Modeling Parameters:} (\textbf{A}) Ablation study with respect to the duration of the experiment in number of trials. Shading denotes 95\% CI derived through bootstrap resampling, see also Table~\ref{tab:classification_performance}. (\textbf{B}) Analysis of model sensitivity to the number of bootstrapped trials in each constructed set. Error bars denote 95\% CI. (\textbf{C}) Analysis of model sensitivity to the number of subject-wise trial sets used in training. $N$ refers to the number of sets created separately for each sentiment category. The evaluation setup was fixed to 200 sets per subject and sentiment category. Shading denotes 95\% CI. (\textbf{D}) Group-wise fixation density maps computed for all \textbf{negative} sentences, from 100\,ms before response to the moment of the response, and plotted as a histogram in projected screen dimensions. We applied 20\% zoom to the image center for better visualization.}
    \label{fig:interp2}
\end{figure*}

Given the potential of integrating eye-tracking markers in screening procedures for depression and suicidality, here we examine the effect of the experiment duration, essentially how model performance is affected when reducing the number of experimental trials. We also conduct a sensitivity analysis on two key model parameters, namely the number of trials per bootstrapped set and the number of subject-wise sets created. Figure~\ref{fig:interp2}A-C contains the quantitative results of this analysis in terms of AUC for both \textit{response}- and \textit{reading}-models.

We observe that both models do not experience severe performance regressions when constraining to the first half of the experiment (Figure~\ref{fig:interp2}A), which suggests that including the second half, where trials were just repeated for a second time (see \textit{Methods}) offers diminishing returns. Notably, the \textit{reading}-model scored higher in the first 50\% of the experiment (0.643 AUC, 95\% CI 0.610--0.677), possibly explained by inhibitory effects of reading the same trials in the second half. In both cases, including the full set of sentences is significant for retaining high performance (two-sample t-tests $p<0.001$). Panel B further illustrates that our trial bootstrapping strategy significantly improves model performance for the response-model (two-sample t-test $p<0.001$) and moderately for the \textit{reading}-model ($p=0.119$). However, performance does not scale with further increasing the size of the sets, as input variability is effectively reduced. Lastly, Panel C demonstrates the robustness of the proposed model, in that performance is retained even with 50 input trial sets per individual (while keeping set size fixed to 30 trials). Performance naturally drops significantly when considering only 1 or 10 sets, which cannot cover the full set of 296 experimental trials.

\section*{Discussion}

This study demonstrates that combining eye-tracking data and deep learning could aid in the identification and better interpretation of mental health disorders. By leveraging fine-grained gaze measures, particularly during the subject-response phase of the experiment, we build on prior work identifying cognitive markers of depression and suicidality~\cite{joormann2010emotion}. With an AUC of 0.793, the classifier demonstrates confident discriminatory power, suggesting that eye movements can serve as an objective measure in systematic assessment tools. Still, the observed specificity (0.674 at the 0.5 threshold) indicates a moderate rate of false positives, with some control participants incorrectly classified as clinical. This limitation is partly due to the class imbalance in our training data, which included a larger proportion of clinical subjects than is typical in the general population. Nonetheless, prioritizing sensitivity may be advantageous for mental health screening, as it reduces the likelihood of missing high-risk individuals who would benefit from further clinical evaluation.

The fact that the model consistently assigned higher scores to suicidal (S) than depressed (D) participants further suggests an implicit sensitivity to suicide-specific features. This is further supported by the significant alignment between model scores and measures of anxiety and suicidality (GAD-7, SIS). Our results further reflect an influence of stimulus valence: classification performance was stronger when the model was trained on negative sentences, consistent with prior studies showing that individuals with depression and suicidality exhibit heightened cognitive and emotional responses to negatively valenced stimuli~\cite{Maalouf2012Bias, joormann2010emotion}. Moreover, performance reductions observed when evaluating the \textit{reading}-model highlight the importance of capturing active, decision-making phases of behavior, which likely reflect attentional biases and maladaptive emotion regulation strategies characteristic of depressed populations~\cite{joormann2010emotion}.

Within the specific context of our experiment, these findings may also suggest that attentional biases preceding the subject's response are shaped not only by abstract cognitive markers but also by the tendency to agree or disagree with each sentence. Prior research~\cite{cavanagh2014eye,smith2019gaze} has demonstrated that gaze patterns reflect and even influence value-based choices when individuals decide between competing alternatives. In our study, both saliency analysis and classifier weights point to a strong temporal focus near the response moment, particularly for negative trials where group differences in endorsement rates are expected. To test whether participants visually anchor on the response option (e.g., ``agree" or ``disagree") they intend to select, we trained a model using only the ratio of time spent fixating on each of those prompts as input. This approach yielded an AUC of 0.694 (95\% CI 0.665, 0.727), significantly above chance but notably lower than our proposed framework on the CvDS task (two-sample t-test $p<0.001$).
Additionally, we trained a logistic regression classifier with only the participants' responses as input, and derived an AUC of 0.940 on the same task, clearly indicating that the \textit{response}-model would attempt to decode the response-group alignment. Figure~\ref{fig:interp2}D depicts the fixations made to the two options shortly before responding to negative sentences, averaged over participants in each group. The histograms point to a clear preference from the control group for the ``Disagree" option, whereas selection bias is less pronounced for the clinical groups, with S further indicating less fixations on the prompts.

Our analysis thus reveals that, while fixation on the final response carries moderate predictive information, our eye-movement model captures additional gaze dynamics and covert attentional biases that may not always be evident in overt behavioral choices. Notably, suicidal participants exhibited more discriminative eye movement patterns approximately 500\,ms prior to their response—rather than just preceding it—suggesting that differences related to suicidality may arise during an earlier, sustained phase of cognitive processing (Figure~\ref{fig:interp}A, right). A similar pattern of temporally intermediate activation is observed in the attribution maps of the \textit{reading}-model for the same group (Figure~\ref{fig:plots-read}C). This finding aligns with theoretical accounts of suicidality that emphasize heightened psychological pain as a central mechanism underlying suicidal ideation~\cite{shneidman1993suicide}. Future work could examine whether these temporal signatures extend to other objective indicators of cognitive processing, such as neurophysiological measures of brain activity.

From a technical perspective, our deep learning approach proved effective not only in achieving high discriminatory performance but also in revealing subtle behavioral signatures embedded in eye movement dynamics that may elude traditional analytic methods. The model’s capacity to highlight such signatures opens avenues for new research hypotheses and experimental analyses of the data at hand. Nonetheless, deep models also have limitations: they are data-hungry, sensitive to dataset biases, and often struggle to generalize to new subjects. In our study, we addressed key challenges of short, event-based eye movement analysis by adopting a within-subject trial bootstrapping strategy (Figure~\ref{fig:methods}E) 
 that resulted in significant performance gains (Figure~\ref{fig:interp2}B) as well as a state-of-the-art architecture to model intra-trial and inter-trial variations. Exposing the model to multiple trials per subject ensured consistency, while randomizing trial order promoted regularization and reduced overfitting. Future studies in computational psychiatry should leverage larger, more diverse populations and more naturalistic environments to enhance reliability.
\section*{Methods}

\subsection*{Recruitment and Group Assignment}

A cohort of 160 young adults (99 females) between 18 and 25 years old was recruited in the city of Los Angeles for this study. Recruitment was done through fliers and in collaboration with the Departments of Psychology and Psychiatry and both the Counseling and Mental Health as well as the Psychiatry and Behavioral Health Services of the University of Southern California. All participants were fluent in English, which was used for the experiment, and completed a series of questionnaires to determine their eligibility for the study as well as their stress, depression, and suicide risk levels. These included the Patient Health Questionnaire (PHQ-9~\cite{spitzer1999validation}) and the Suicidal Ideation Scale (SIS~\cite{rudd1989prevalence}). Based on those two scores, they were grouped into three groups: controls (C, PHQ-9 $<5$), depressed (D, PHQ-9 $>9$ and SIS $\leq 16$), or suicidal (S, PHQ-9 $>9$ and SIS $>16$). Participants also completed the GAD-7~\cite{spitzer2006brief} questionnaire, related to anxiety.

The questionnaires were administered once at the recruitment phase and once after the completion of the experiment. 13 participants with unstable PHQ-9 scores between the time of recruitment and the time of data acquisition (i.e., having a shift that caused them to change groups or fall into an ineligible range) as well as 1 participant that did not complete the experiment were excluded from the present analysis. Out of the remaining 146 participants, 49 were assigned to group C, 47 to group D, and 50 to group S. Further, all subjects were excluded who had the following characteristics: (1) current or previous diagnosis of neurological and psychiatric disorders including Schizophrenia, Bipolar Disorder, Epilepsy, Brain Cancer, and Stroke, (2) color blindness, (3) having learned English after the age of 7, and (4) a PHQ-9 score between 5 and 9.

\subsection*{Data Collection}

The participants underwent a visual sentence viewing task during which EEG, eye tracking, and physiological signals were simultaneously collected. Here we only consider the eye-tracking portion of the data. Eye movements were continuously recorded using the SR Research EyeLink 1000 Plus at a sampling rate of 500 Hz (Figure~\ref{fig:methods}B). Participants were shown sentences consisting of 5 to 11 words. The sentences were presented word by word with an inter-word interval IWI$=300$\,ms (blank screen). Each word was displayed for 300\,ms, except the last word, which was displayed for 600\,ms (Figure~\ref{fig:methods}A). After offsetting the final word and an additional 300\,ms interval, participants were instructed to press one of two available buttons to indicate whether they agree (congruence) or disagree (incongruence) with the sentence. The prompt remained on the screen until the button was pressed or for up to two seconds; if no response was made within this time, the trial was marked as missed. The whole recording session lasted about 55 minutes.

Each sentence was presented twice, once at the first and once at the second half of the experiment, in a random place, starting with a central fixation cross. There were 160 distinct sentences in total, generated as 80 pairs; within each pair, the sentences were identical except for the final word, which yielded opposite overall sentiment; for example: ``\textit{My mental health is not sound}" and ``\textit{My mental health is not problematic}". The negative sentences were assumed to be congruent with depression or suicidality, whereas the positive/neutral sentences were not (Figure~\ref{fig:methods}F). Only the last word resolved the relationship between a sentence’s meaning and a participant's experience and reflections; we thus manually excluded 12 of the sentences from further analysis as behavioral outliers based on participants' average responses. Final words themselves were balanced for valence, including positive, neutral, and negative words, across the two types of sentences.

\subsection*{Data Pre-processing}

Eye-tracking markers for fixations, saccades, and blinks were obtained from the raw recording. This information was used to denoise the signal as follows. First, gaze coordinates were bound to the spatial constraints of the computer display (1920×1080 pixels). Artifacts—including blinks and saccadic outliers (defined as velocities exceeding 3 standard deviations from the mean)—were identified and replaced via linear interpolation. Initially in pixel coordinates, the gaze time series were normalized to a coordinate system centered on the screen midpoint (0,0), where stimuli appeared, with edges mapped to (±1, ±1). At this stage, 20 participants were excluded from further analysis because their eye-tracking recording was either missing~(7) or severely distorted~(13). Finally, the gaze recordings from the remaining 126 participants were downsampled to 250 Hz.

The eye-tracking trials were segmented using two strategies, corresponding to two different phases of participant response. The first strategy focused on the response associated with stimulus viewing. It included the time interval from 500\,ms before until 900\,ms after the onset of the critical final word of the trial sentence (Figure~\ref{fig:methods}C--up). The upper bound was selected because at this point the participants were prompted on-screen to state if they agree or disagree with the presented sentence. The second strategy focused on the response associated with decision making (button press). It included the time interval from 1200\,ms before to the exact time of the button press (Figure~\ref{fig:methods}C--down). Here, the lower bound was defined as the mode of the distribution of the response times with respect to final-word onset (Figure~\ref{fig:methods}D).

\subsection*{Bootstrapping Augmentation}

We train the deep learning models using the final 296 trials for each of the 126 participants, that is, about 37,000 trials in total. However, the information to be gained from a single trial input is constrained. To tackle this limitation and augment the training process, we instead input sets of 30 trials (10\% of the total) to the model, each set constructed by sampling with replacement from all 296 trials of each participant and sentiment category separately (Figure~\ref{fig:methods}E). To save on computational resources, we construct 200 such sets for each subject and sentiment category (400 in total), prior to model training. Each contains a random subset of 30 trials, organized in a 2D format for x and y traces separately (Figure~\ref{fig:methods}E--right).

\subsection*{Model Architecture}

The model architecture used to process the eye-tracking data is schematically shown in Figure~\ref{fig:methods}F. It consists of two backbone components, a late fusion mechanism and a classifier. We used one backbone component to process the trials of negative valence and a separate one for the remaining trials. These consist of a stack of 2D convolutional layers designed to capture intra-trial and inter-trial characteristics. This method was introduced in TimesNet~\cite{timesnet}, a state-of-the-art model for time-series analysis. We used the same configuration as TimesNet and applied a series of two Inception layers~\cite{inception} to capture both local and global dependencies effectively, i.e. by applying multiple 2D convolutional filters of different sizes in parallel. Their output is then averaged along the trials' dimension and $(x, y)$ channels are concatenated. Subsequently, the latent feature vectors of each backbone are fused, and the output is sent to a fully connected classifier. The model outputs a scalar probability estimate that the input sample belongs to the positive (typically, the clinical) class.

\subsection*{Training and Validation Protocol}

In all deep learning experiments, we used a nested cross-validation scheme with 5 outer and 10 inner folds. Each outer fold consisted of a non-overlapping set of participants, so that all participants were in exactly one of the folds and no test participant was parsed in training. Folds were themselves stratified based on gender and clinical group. Each inner fold was determined by a random seed and was used to tune the parameters of each model as well as for early stopping. To assess model performance, we report the area under the ROC curve (AUC) as well as sensitivity and specificity scores for the standard classification threshold of 0.5. All models were trained using class weights for the objective function that are inversely proportional to class distribution in the training set. The trainer class used a batch size of 64 samples, a learning rate of 5e-4 and early stopping after 5 epochs of non-decreasing cross-entropy loss in the validation set. During inference, the model outputs a prediction for all 200 samples per participant. Then, the percentage of positive predictions (i.e., $P_{=1}\geq 0.5$) is used as the final participant-level score (Figure~\ref{fig:methods}F--right).

Confidence intervals were derived empirically through bootstrap resampling with 1000 iterations, considering the predictions of the entire dataset. To establish an empirical null distribution, we performed permutation testing by training identical models against randomly permuted labels, followed by the same resampling procedure. Statistical significance was quantified by counting the proportion of null-distribution bootstrap means that exceeded the bottom 95\% CI of model performance, which effectively let us determine p-values down to 0.001. We used False Detection Rate (FDR) to correct for multiple comparisons.

\section*{Data Availability}
The full set of anonymized eye-tracking recordings and annotations are available upon request to shri@usc.edu.

\section*{Code Availability}
All presented analyses were implemented in Python (v12.0). All models were trained using two NVIDIA RTX 6000 GPUs within a secure server. The associated code is available upon request. 

\bibliography{References}

\section*{Acknowledgments}
This study was sponsored by the Defense Advanced Research Projects Agency (DARPA) under cooperative agreement No. N660012324006. The content of the information does not necessarily reflect the position or the policy of the Government, and no official endorsement should be inferred.

\section*{Author Contributions}
K.A. contributed to the conception, design, analysis, and evaluation of the computational framework for eye movement analysis. K.A. and M.M. pre-processed the eye-tracking recordings. W.J., C.M., M.H. and T.M. contributed to the experimental design and acquisition of the data. D.B., A.H., R.C., T.M., I.B., K.L., R.L., and S.N. contributed to the conception and design of the experimental procedure and data collection protocol. W.J., A.K., S.K., M.M., E.K., and D.B. contributed to the data analysis and interpretation of model results. R.L., I.B., E.K., D.B. and S.N. provided scientific oversight and supervision throughout the study implementation. All authors contributed to the writing of the manuscript and the figures.

\section*{Competing Interests}
The authors declare no competing interests.

\newpage
\section*{Supplementary Figures}

\end{document}